\documentclass[10pt,twocolumn,letterpaper]{article}

\usepackage{iccv}
\usepackage{times}
\usepackage{epsfig}
\usepackage{graphicx}
\usepackage{amsmath}
\usepackage{amssymb}
\usepackage{caption}
\usepackage{subcaption}
\usepackage[x11names,dvipsnames,table]{xcolor} 
\usepackage{multirow}
\usepackage{array}
\newcolumntype{L}[1]{>{\raggedright\let\newline\\\arraybackslash\hspace{0pt}}m{#1}}
\newcolumntype{C}[1]{>{\centering\let\newline\\\arraybackslash\hspace{0pt}}m{#1}}
\newcolumntype{R}[1]{>{\raggedleft\let\newline\\\arraybackslash\hspace{0pt}}m{#1}}

\usepackage[pagebackref=true,breaklinks=true,letterpaper=true,colorlinks,bookmarks=false]{hyperref}

\iccvfinalcopy 
\def\iccvPaperID{2131} 

\begin{document}

\title{Phase Collaborative Network for Two-Phase Medical Imaging Segmentation}

\author{Huangjie Zheng\textsuperscript{1,3}\thanks{This work is done when this author was working in SJTU and JHU}, Lingxi Xie\textsuperscript{2,4}, Tianwei Ni\textsuperscript{3}, Ya Zhang\textsuperscript{1}, Yan-Feng Wang\textsuperscript{1}, Qi Tian\textsuperscript{4}\\ Elliot K. Fishman\textsuperscript{5}, Alan L. Yuille\textsuperscript{2}\\
\textsuperscript{1}Shanghai Jiao Tong University\quad\textsuperscript{2}The Johns Hopkins University \quad \textsuperscript{3}University of Texas at Austin\\
\textsuperscript{4}Carnegie Mellon University\quad
\textsuperscript{5}Huawei Noah’s Ark Lab\quad 
\textsuperscript{6}The Johns Hopkins School of Medicine\\
{\tt\small \{zhj865265, ya\_zhang, wangyanfeng\}@sjtu.edu.cn}\quad{\tt\small \{198808xc, twni2016, alan.l.yuille\}@gmail.com}\\
{\tt\small huangjie.zheng@utexas.edu} \quad {\tt\small tian.qi1@huawei.com} \quad {\tt\small efishman@jhmi.edu}\\
}

\maketitle

\begin{abstract}
In real-world practice, medical images acquired in different phases possess complementary information, {\em e.g.}, radiologists often refer to both {\em arterial} and {\em venous} scans in order to make the diagnosis. However, in medical image analysis, fusing prediction from two phases is often difficult, because (i) there is a domain gap between two phases, and (ii) the semantic labels are not pixel-wise corresponded even for images scanned from the same patient. This paper studies organ segmentation in two-phase CT scans. We propose Phase Collaborative Network (PCN), an end-to-end framework that contains both generative and discriminative modules. PCN can be mathematically explained to formulate phase-to-phase and data-to-label relations jointly. Experiments are performed on a two-phase CT dataset, on which PCN outperforms the baselines working with one-phase data by a large margin, and we empirically verify that the gain comes from inter-phase collaboration. Besides, PCN transfers well to two public single-phase datasets, demonstrating its potential applications. 
\end{abstract}

\section{Introduction}
\label{Introduction}

Semantic segmentation is one of the fundamental problems in computer vision which implies a wide range of applications. Recent years, with the rapid development of deep learning~\cite{lecun2015deep,krizhevsky2012imagenet,simonyan2014very,he2016deep}, researchers have designed powerful segmentation models~\cite{long2015fully,chen2014semantic,chen2016semantic,chen2018deeplab,dai2016instance} which are mostly equipped with an encoder-decoder architecture. These models have achieved success in various image domains, including medical image analysis, in particular organ and soft-tissue segmentation, which forms an important prerequisite of computer-assisted diagnosis~\cite{ronneberger2015u,roth2015deeporgan,zhou2017fixed,yu2018recurrent}.

Medical images can appear in more than one {\em phases}, each of which corresponds to a specific way of data sampling and scanning. It has been well acknowledged that incorporating multi-phase information improves visual recognition~\cite{5238600,YANG20141559}. Nevertheless, there have fewer studies on this problem. There are two possible reasons -- one of them lies in the lack of multi-phase training data, and the other refers to the difficulty in aligning multi-phase data and digging complementary information out from them.

In this paper, we study this issue in the field of CT scans, for which we construct a large-scale dataset of $200$ patients. For each case, two 3D volumes were collected from the {\em arterial} and {\em venous} phases, and the radiologists in our team manually annotated several abdominal targets, including organs and blood vessels. This is to say, in our dataset, each sample is composed of two paired images from {\em arterial} and {\em venous} phase, respectively. Note that the images scanned at the same position can be largely different, due to the difference of radiation in scanning. Plus, although they are scanned from the same patient, the organs and vessels are not corresponded in pixel level due to their motion in the human body (see Figure~\ref{Fig:Examples}). This causes huge difficulties in registrations \cite{warfield2004simultaneous,asman2013non} between the two phases. Our goal is to train a model that leverages the information from both  phases in a collaborative way and improves the segmentation, while conventional approaches dealt with data in either phase, but missed the inter-phase connection.


\newcommand{\colwidth}{1.00cm}
\begin{table}
\centering
\setlength{\tabcolsep}{0.04cm}
\resizebox{\linewidth}{!}{
\begin{tabular}{|l||C{\colwidth}|C{\colwidth}|C{\colwidth}|C{\colwidth}|C{\colwidth}|}
\hline
{} & CycleGAN \cite{CycleGAN2017} & CyCADA \cite{Hoffman_cycada2017} & UCDA \cite{dou2018unsupervised}& SIFA \cite{chen2019synergistic} & {\bf PCN} (ours) \\
\hline
Image adaptation     &    \checkmark   &    \checkmark   &     &    \checkmark        & \checkmark \\ 
\hline
Feature adaptation     &        &    \checkmark   & \checkmark    &     \checkmark       & \checkmark \\
\hline
Collaborative learning   & \checkmark  &   &    &    & \checkmark \\
\hline
Unknown label inference &   & \checkmark & \checkmark & \checkmark & \checkmark\\
\hline
\end{tabular}}
\caption{A comparison between our problem setting and that of previous approaches. Our approach stands out with the task of collaborative learning (see texts for details).
\vspace{-0.58cm}
\noindent
}
\label{tab:comparison}
\end{table}

To model the inter-phase relation without the need of inter-phase registration, the problem refers to domain transfer and domain adaptation. However, as shown in Table~\ref{tab:comparison}, our setting is clearly different from that of existing approaches. More specifically, the knowledge from two phases needs to `help each other' during training and testing -- we call it {\bf phase collaboration}. To this end, we propose an end-to-end framework named Phase Collaborative Network (PCN), which formulates the joint distribution of two-phase-image data and their semantic labels. The major contribution of this work lies in decomposing this distribution into two parts, namely, a data-to-label relation and a phase-to-phase relation. In practice, the former term is implemented as a discriminative model ({\em e.g.}, a segmentation network), and the latter one as a generative model ({\em e.g.}, a Generative Adversarial Network \cite{goodfellow2014generative} which can transfer the image style across different phases). We adopt a multi-stage strategy to train PCN, so as to facilitate the two modules to cooperate while guaranteeing the stability of optimization.

We evaluate PCN on two sources of data, including our own two-phase dataset and two public one-phase datasets. In our own data, PCN learns two functions for {\em arterial}-to-{\em venous} and {\em venous}-to-{\em arterial} transfer, respectively, so that the segmentation in each phase can be assisted by another one, and the accuracy is boosted consistently. In particular, when the target is difficult ({\em e.g.}, a small target like the {\em adrenal gland}) or less discriminative under the specific phase ({\em e.g.}, an {\em artery} in the {\em venous} phase), significant accuracy gain is obtained. In public datasets where only the {\em venous} phase is present, PCN takes advantage of the {\em venous}-to-{\em arterial} transfer function learned from the two-phase dataset and generates {\em arterial} data as extra knowledge. This helps PCN outperform existing approaches that use only single-phase information. This demonstrates the potential of PCN, which can be trained at one time with paired two-phase training data, and freely applied to other one-phase scenarios.

The remainder of this paper is organized as follows. Section~\ref{RelatedWork} briefly reviews related work, and Section~\ref{Problem} introduces the problem setting. The core part, Phase Collaborative Network, is described in Section~\ref{Approach}. Experiments are shown in Section~\ref{Experiments}, and the conclusions drawn in Section~\ref{Conclusions}.

\section{Related Work}
\label{RelatedWork}

\textbf{Semantic segmentation} is a critical problem in computer vision. Recently, with the rapid development of deep learning, conventional approaches built upon graph-based algorithms~\cite{ali2007graph} and/or handcrafted local features~\cite{wang2014geodesic} have been replaced by deep neural networks~\cite{lecun2015deep,krizhevsky2012imagenet,simonyan2014very,he2016deep} that can produce higher segmentation accuracy~\cite{long2015fully,chen2018deeplab}. The progress in segmentation has boosted more vision tasks such as video-based segmentation~\cite{perazzi2016benchmark,caelles2017one}, instance segmentation~\cite{pinheiro2015learning,hariharan2014simultaneous,pinheiro2014recurrent,shelhamer2016clockwork,tran2016deep} and 3D segmentation~\cite{huang2016point,qi2017pointnet}.

\textbf{Medical imaging analysis} is an important prerequisite of computer-aided diagnosis (CAD) that can assist human doctors in clinical scenarios. Since medical images contain enormous information such as internal organs, bones, soft tissues and vessels, automatic segmentation plays a fundamental role of further diagnoses~\cite{brosch2016deep,wang2016deep,havaei2017brain,zhou2017deep}. Researchers have designed individualized algorithms in order to capture specific properties of different organs, {\em e.g.}, liver~\cite{ling2008hierarchical,heimann2009comparison}, spleen~\cite{linguraru2010automated}, kidneys~\cite{lin2006computer}, lungs~\cite{hu2001automatic}, pancreas~\cite{chu2013multi,zhou2017fixed,yu2018recurrent}, {\em etc}. Most existing approaches focused on single-phase data, however, human doctors often refer to multi-phase data. To bridge this information gap, researchers attempted to introduce multi-phase information to improve segmentation accuracy~\cite{linguraru2010multi,wolz2013automated}. However, since the internal organs are not still during the scanning process, such algorithms often face the problem of alignment. The intrinsic problem lies in the lack of paired data across phases, which makes it difficult to apply existing algorithms for phase registration~\cite{asman2013non,warfield2004simultaneous,wang2013multi}. This drives us to formulate the relationship between different phases with a generative model.

\textbf{Deep generative models} aim at using a parametric distribution to fit the real data distribution in a unsupervised manner. Modeling this distribution can achieve the goal of data generation. In recent years, generative models like VAE and GAN~\cite{kingma2013auto,goodfellow2014generative} and their variants have become quite popular in both theory and applications. Arterial and venous images are sampled form different distributions, and building a relationship between them refers to the field of domain adaptation, where representative approaches include Pix2Pix~\cite{isola2017image}, CycleGAN~\cite{CycleGAN2017} and UNIT~\cite{liu2017unsupervised}. The basic assumption of these approaches is that image data are composed of `content' and `style', and they focuses on style transfer while the content remains unchanged~\cite{gatys2016image,yi2017dualgan}. In our work, we consider the organ annotations as the content and their appearance in different phases as style. These domain transfer methods were also applied to cross-domain segmentation~\cite{hoffman2016fcns,Hoffman_cycada2017,luc2016semantic}, but their setting is different from ours, in which data and label appear in both phases (there are no `source' and `target' domains) and two labels are related to the same, unobserved distribution.

\section{Problem: Two-Phase Segmentation}
\label{Problem}

We consider semantic segmentation in the context that information comes from two phases, which is a common case of medical data. Mathematically, there are two phases named $\mathcal{A}$ and $\mathcal{V}$: $\mathcal{A}=\{\mathbf{X}_\mathrm{A}\in\mathbb{R}^D,\mathbf{Y}_\mathrm{A}\in\{0,1\}^D\}$ and $\mathcal{V}=\{\mathbf{X}_\mathrm{V}\in\mathbb{R}^D,\mathbf{Y}_\mathrm{V}\in\{0,1\}^D\}$, where $\mathbf{X}$ and $\mathbf{Y}$ denote the image and the annotation of dimension $D$; the subscripts, $\mathrm{A}$ and $\mathrm{V}$, denote {\em arterial} and {\em venous}. Here we use the subscripts $\mathrm{A}$ and $\mathrm{V}$ to denote two popular phases named {\em arterial} and {\em venous} in CT scans, which are the stage our approach is working on. 

\begin{figure}[!t]
\centering
\begin{subfigure}{.23\textwidth}
\centering
\includegraphics[width=.9\textwidth]{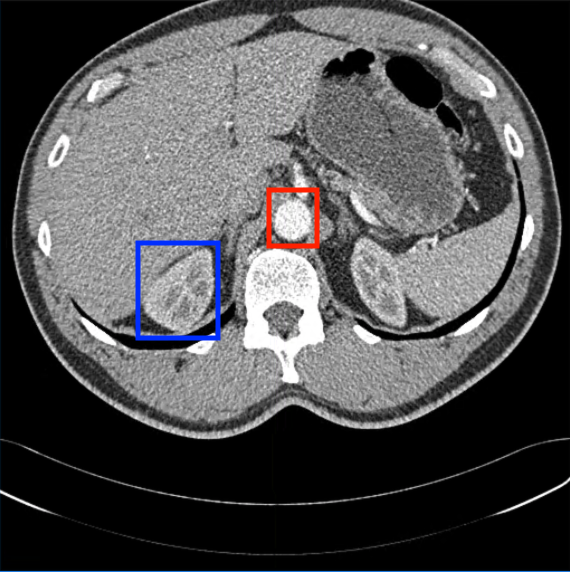}
\caption{an arterial-phase CT scan}\label{fig:arterial}
\end{subfigure}
\hfill
\begin{subfigure}{.23\textwidth}
\centering
\includegraphics[width=.9\textwidth]{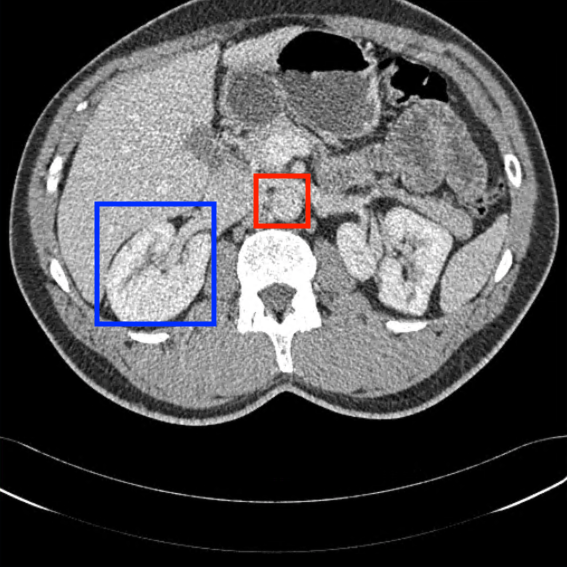}
\caption{a venous-phase CT scan}\label{fig:venous}
\end{subfigure}
\caption{(\textit{Best viewed in color}) An illustration of the differences between CT scans from the {\em arterial} (\textbf{left}) and {\em venous} (\textbf{right}) phases. The {\em artery} (marked in \textcolor{red}{red}) has a larger intensity in the {\em arterial} phase while the {\em kidney} (marked in \textcolor{blue}{blue}) other organs have larger intensities in the {\em venous} phase. These differences are mainly due to the different properties (in radiology) of these targets.
\vspace{-0.35cm}
\noindent
}
\label{Fig:Examples}
\end{figure}

A pair of examples of {\em arterial} and {\em venous} scans are shown in Figure~\ref{Fig:Examples}. Both scans are performed on the same patient, but the {\em arterial} scan happens $35$ seconds prior to the {\em venous} scan, when the radiation level is higher and thus {\em artery}-related organs and blood vessels are of higher intensity than those in the {\em venous} scan (in opposite, {\em vein}-related targets are of higher intensity in the {\em venous} scan). This means that $\mathbf{X}_\mathrm{A}$ and $\mathbf{X}_\mathrm{V}$ are sampled from two distributions of different appearances, but $\mathbf{Y}_\mathrm{A}$ and $\mathbf{Y}_\mathrm{V}$ should be from the same distribution, {\em i.e.}, the internal structure of the same person does not change. However, since organs and blood vessels of a living person are not still during the scanning process, the observed $\mathbf{Y}_\mathrm{A}$ and $\mathbf{Y}_\mathrm{V}$ are actually different from each other. In our case, the radiologists in our team annotated the two phases individually, which further increases the extent of nuance. In overall, $\mathbf{Y}_\mathrm{A}$ and $\mathbf{Y}_\mathrm{V}$ can be considered sampled from the same distribution centered at $\mathbf{Y}^\star$, denoted as $\mathcal{N}\!\left(\mathbf{Y}^\star,\boldsymbol{\Sigma}\right)$. This inspires the labels can be predict with information from both phases. The goal is thus to train a segmentation model for ${\mathbf{Y}_\mathrm{A}}={\mathbf{f}\!\left(\mathbf{X}_\mathrm{A},\mathbf{X}_\mathrm{V};\boldsymbol{\theta}\right)}$ or ${\mathbf{Y}_\mathrm{V}}={\mathbf{f}\!\left(\mathbf{X}_\mathrm{A},\mathbf{X}_\mathrm{V};\boldsymbol{\theta}\right)}$, during which both {\em arterial} and {\em venous} training data are used, as well as the information in these two phases are considered in a collaborative way. 

\section{Approach: Phase Collaborative Network}
\label{Approach}

\subsection{Phase Collaboration}
\label{Approach:Collaboration}
The main idea of phase collaboration is to build relation between two phases so that both of them benefit from complementary information. Therefore, two phases need to be considered simultaneously, unlike previous approaches, such as~\cite{Hoffman_cycada2017} that was trained on one phase and transferred to another. The intuition is confirmed by an entropy inequality~\cite{information_inequality}, {\em i.e.}, ${\mathcal{H}\!\left(\mathbf{X}_\mathrm{A},\mathbf{X}_\mathrm{V}\right)}\geqslant{\max\left\{\mathcal{H}\!\left(\mathbf{X}_\mathrm{A}\right),\mathcal{H}\!\left(\mathbf{X}_\mathrm{V}\right)\right\}}$, where $\mathcal{H}\!\left(\cdot\right)$ indicates the Shannon entropy. Mathematically, our goal is to learn a model $\mathbf{f}\!\left(\mathbf{X}_\mathrm{A},\mathbf{X}_\mathrm{V};\boldsymbol{\theta}\right)$ which minimizes $\left\|\mathbf{f}\!\left(\mathbf{X}_\mathrm{A},\mathbf{X}_\mathrm{V};\boldsymbol{\theta}\right)-\mathbf{Y}_\mathrm{A}\right\|+\left\|\mathbf{f}\!\left(\mathbf{X}_\mathrm{A},\mathbf{X}_\mathrm{V};\boldsymbol{\theta}\right)-\mathbf{Y}_\mathrm{V}\right\|$. Due to the symmetry of these two terms, we simply discuss the first term, related to $\mathrm{A}$, in the following parts.

The objective of minimizing $\left\|\mathbf{f}\!\left(\mathbf{X}_\mathrm{A},\mathbf{X}_\mathrm{V};\boldsymbol{\theta}\right)-\mathbf{Y}_\mathrm{A}\right\|$ is to integrate information from both phases, which is to say, to model the relations between $\mathbf{X}_\mathrm{A}$, $\mathbf{X}_\mathrm{V}$ and $\mathbf{Y}_\mathrm{A}$. Since $\mathbf{X}_\mathrm{A}$ and $\mathbf{X}_\mathrm{V}$ are not acquired in pairs, it is difficult to directly model their relation using methods such as registration \cite{warfield2004simultaneous,asman2013non}, as shown in Figure \ref{Fig:Examples}.  

\begin{figure}[!t]
 \centering
\includegraphics[width=\linewidth]{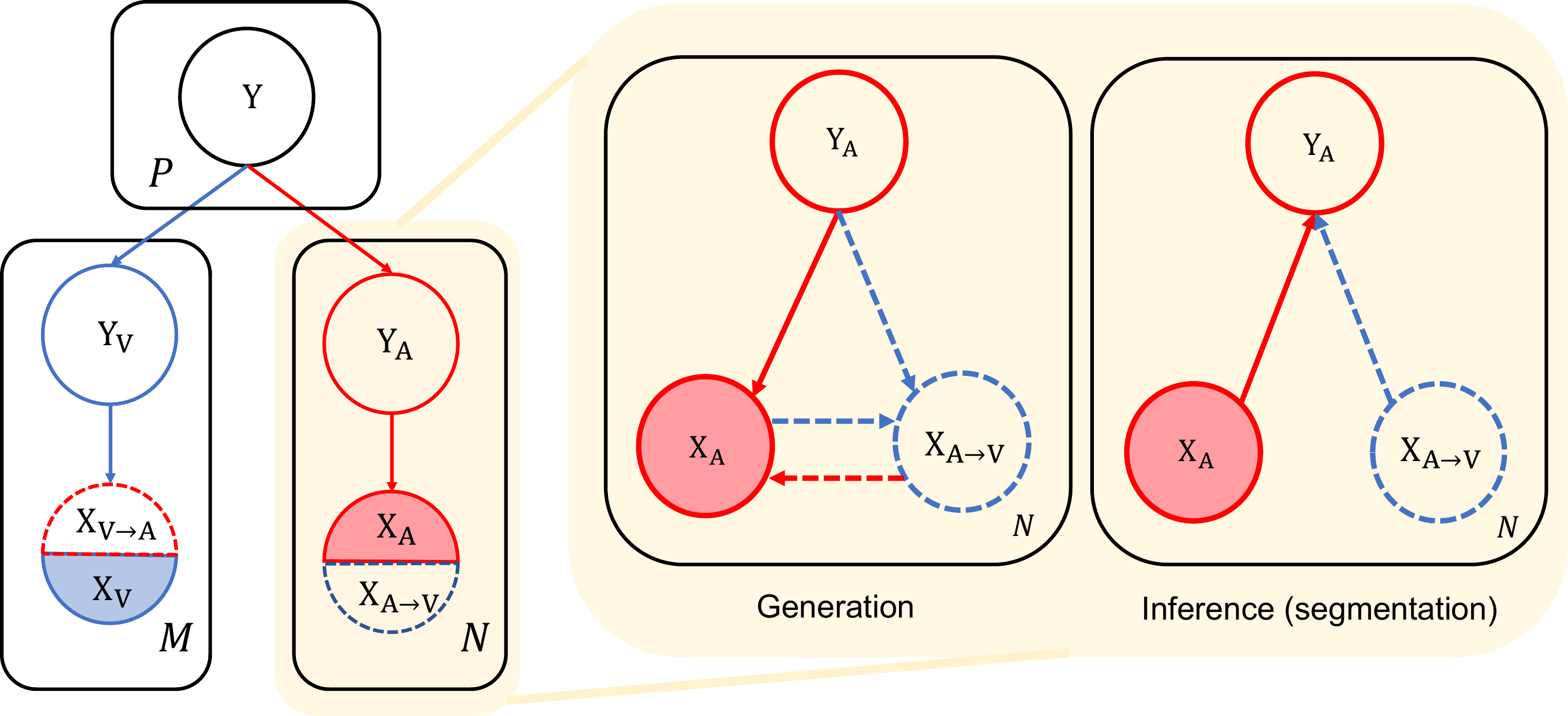}
\caption{(\textit{Best viewed in color}) A graphical representation in a two-phase scenario. The {\bf left} figure shows that $\mathbf{Y}_\mathrm{A}$ and $\mathbf{Y}_\mathrm{V}$, the labels in two phases, are sampled from the same distribution $p\!\left(\mathbf{Y}\right)$, while the distributions of $\mathbf{X}_\mathrm{A}$ and $\mathbf{X}_\mathrm{V}$ are different. The right figure shows the generation and inference processes in the {\em arterial} phase. During generation, each fake {\em venous} image, $\mathbf{X}_{\mathrm{A}\rightarrow\mathrm{V}}$, is generated with the semantic label $\mathbf{Y}_\mathrm{A}$ and the phase-specific style elements. The inference is the process where the semantic label $\mathbf{Y}_\mathrm{A}$ can be inferred from both $\mathbf{X}_\mathrm{A}$ and $\mathbf{X}_{\mathrm{A}\rightarrow\mathrm{V}}$. Unobserved data are marked with dashed borders, and the process related to phase $\mathrm{A}$ (\textit{resp.} $\mathrm{V}$) is marked in \textcolor{red}{red} (\textit{resp.} \textcolor{blue}{blue}). $P$ marks the number of patients; $M$ and $N$ mark the number of samples in venous and arterial phases, respectively.
\vspace{-0.35cm}
\noindent
}
\label{Fig:GraphicalModel}
\end{figure}

We propose a decomposed relation to better model $\mathbf{X}_\mathrm{A}$, $\mathbf{X}_\mathrm{V}$ and $\mathbf{Y}_\mathrm{A}$, and apply a probabilistic graphical model~\cite{koller2009probabilistic} in Figure~\ref{Fig:GraphicalModel} to help understand these relations:
\begin{align} \nonumber
\begin{split}
\!\!\!{p_\mathrm{data}\!\left(\mathbf{X}_\mathrm{A},\mathbf{X}_\mathrm{V},\mathbf{Y}_\mathrm{A}\right)}\propto{\underbrace{p_\mathrm{data}\!\left(\mathbf{Y}_\mathrm{A}\!\mid\!\mathbf{X}_\mathrm{A}\right)}_\text{data-to-label}\cdot\underbrace{p_\mathrm{data}\!\left(\mathbf{X}_\mathrm{V}\!\mid\!\mathbf{X}_\mathrm{A},\mathbf{Y}_\mathrm{A}\right)}_\text{phase-to-phase}}
\end{split}
\end{align}
where the first term, $p_\mathrm{data}\!\left(\mathbf{Y}_\mathrm{A}\mid\mathbf{X}_\mathrm{A}\right)$, models the relation between the paired data and labels in phase $\mathrm{A}$; while the second term, $p_\mathrm{data}\!\left(\mathbf{X}_\mathrm{V}|\mathbf{X}_\mathrm{A},\mathbf{Y}_\mathrm{A}\right)$, models the relation between phase $\mathrm{A}$ and phase $\mathrm{V}$, respectively.   

As shown, the data-to-label relation indicates that labels can be directly inferred with data acquired in phase $\mathrm{A}$ with segmentation models $\mathbf{f}_{\mathrm{A}}$, where $\mathbf{Y}_\mathrm{A}=\mathbf{f}_{\mathrm{A}}\!\left(\mathbf{X}_\mathrm{A}\right)$. The phase-to-phase relation indicates data acquired in phase $\mathrm{V}$ can be inferred with the knowledge acquired in phase $\mathrm{A}$. Here, we suppose there are samples $\mathbf{X}_{\mathrm{A}\rightarrow\mathrm{V}} \sim p_\mathrm{data}\!\left(\mathbf{X}_\mathrm{V}\!\mid\!\mathbf{X}_\mathrm{A},\mathbf{Y}_\mathrm{A}\right)$ that are unobserved due to the CT scanning mechanism. The unobserved data can be modeled with generators, $\mathbf{G}_{\mathrm{A}\rightarrow\mathrm{V}}$ so that $\mathbf{X}_{\mathrm{A}\rightarrow\mathrm{V}}=\mathbf{G}_{\mathrm{A}\rightarrow\mathrm{V}}\!\left(\mathbf{X}_\mathrm{A},\mathbf{Y}_\mathrm{A}\right)$. $\mathbf{X}_{\mathrm{A}\rightarrow\mathrm{V}}$ represents samples that correspond to the semantic labels $\mathbf{Y}_{\mathrm{A}}$, but with phase-specific styles. Intuitively, for a segmentation model trained with data and labels acquired in phase $\mathrm{V}$, $\mathbf{f}_\mathrm{V}$, we should have $\mathbf{Y}_\mathrm{A}=\mathbf{f}_{\mathrm{V}}\!\left(\mathbf{X}_{\mathrm{A}\rightarrow\mathrm{V}}\right)$. 

\subsection{Formulation}
\label{Approach:Formulation}

\begin{figure*}[tb]
 \centering
 \includegraphics[width=0.95\linewidth]{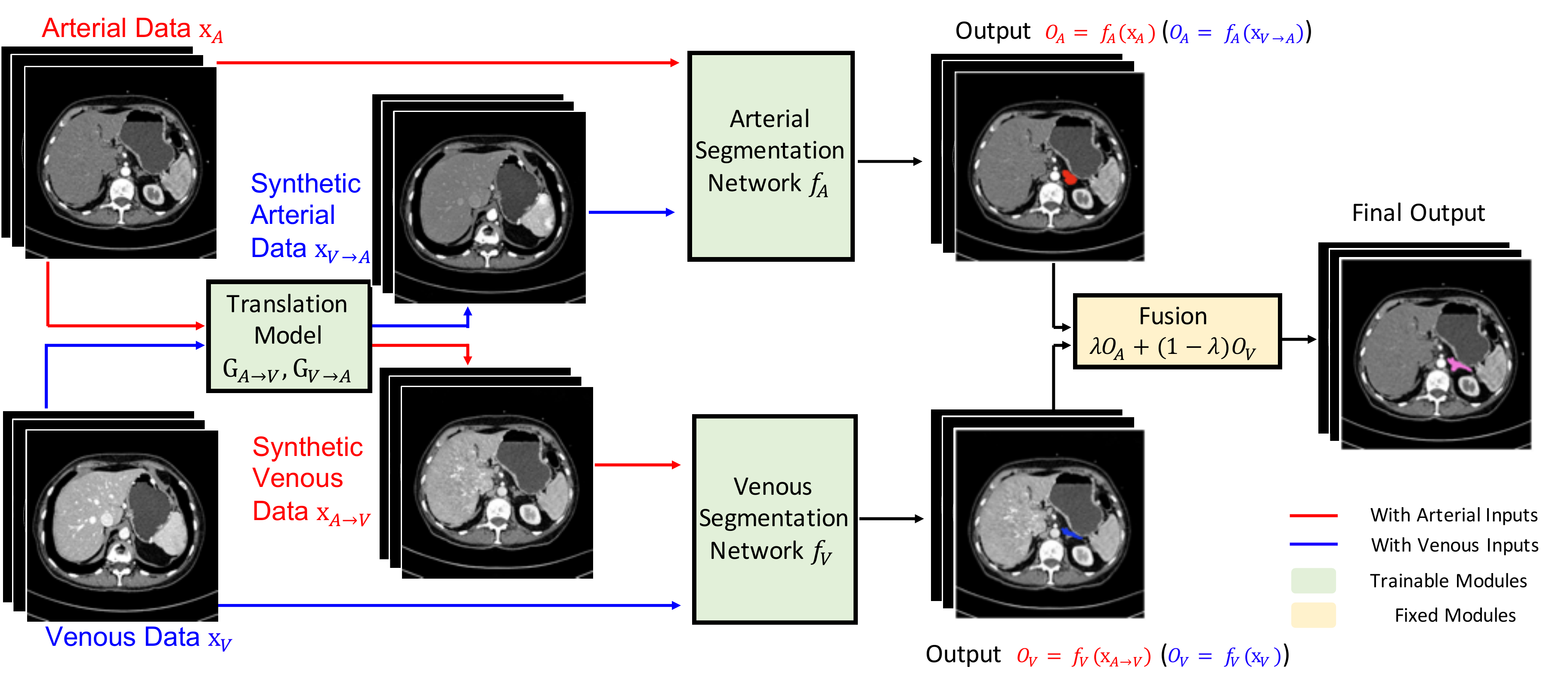}
\caption{(\textit{Best viewed in color}) Illustration of the entire framework of PCN. Image examples are sampled from our two-phase medical segmentation dataset (see Section~\ref{Experiments:Settings}). The learnable modules include two segmentation networks and a bidirectional generator. During training, there are three loss functions to compute, namely the {\em arterial}/{\em venous} segmentation losses and the generator loss. \textcolor{red}{Red} and \textcolor{blue}{blue} arrows indicate information propagation in the {\em arterial} and {\em venous} phases, respectively.}
\label{Fig:Framework}
\end{figure*}

This subsection provides the formulation from the perspective of loss functions. We start with modeling both data-to-label and phase-to-phase relations using deep networks, and then combine them into the overall objective.

\vspace{0.2cm}
\noindent
$\bullet$\quad{\bf Data-to-Label Relation}

\noindent
Data-to-Label Relation can be directly modeled with segmentation models. Given an image-label pair $\left(\mathbf{X}_\mathrm{A},\mathbf{Y}_\mathrm{A}\right)$, the loss function of segmentation model $\mathbf{f}_\mathrm{A}:\mathbf{X}_\mathrm{A}\rightarrow\mathbf{Y}_\mathrm{A}$ is defined as: 
\begin{align}
\label{Eqn:SegmentationLoss}
\begin{split}
\mathcal{L}_\text{D2L}\!\left(\mathbf{X}_\mathrm{A},\mathbf{Y}_\mathrm{A};\mathbf{f}_\mathrm{A}\right) &=\!\left\|\mathbf{f}_\mathrm{A}\!\left(\mathbf{X}_\mathrm{A};\boldsymbol{\theta}_\mathrm{A}\right)-\mathbf{Y}_\mathrm{A}\right\| \\
\end{split}
\end{align}
where $\boldsymbol{\theta}_\mathrm{A}$ indicates the parameter of the segmentation model $\mathbf{f}_\mathrm{A}$. Note that the Data-to-Label relation does not only exist in phase $\mathrm{A}$. In phase $\mathrm{V}$, we can also model this relation with another segmentation model $\mathbf{f}_\mathrm{V}:\mathbf{X}_\mathrm{V}\rightarrow\mathbf{Y}_\mathrm{V}$. Plus, here we apply norm distance, while in other cases this metric can also be replaced by other  segmentation criterion.

\vspace{0.2cm}
\noindent
$\bullet$\quad{\bf Phase-to-Phase Relation}

\noindent As shown in Figure~\ref{Fig:GraphicalModel}, to model the Phase-to-Phase relation, we suppose there exists unobserved data $\mathbf{X}_{\mathrm{A}\rightarrow\mathrm{V}} \sim p_\mathrm{data}\!\left(\mathbf{X}_\mathrm{V}\!\mid\!\mathbf{X}_\mathrm{A},\mathbf{Y}_\mathrm{A}\right)$. We further leverage an intuition that $\mathbf{X}_\mathrm{A}$ and $\mathbf{X}_{\mathrm{A}\rightarrow\mathrm{V}}$ can be transferred to each other and correspond to the same label, \emph{i.e.} $\mathbf{Y}_\mathrm{A}=\mathbf{f}_{\mathrm{A}}\!\left(\mathbf{X}_{\mathrm{A}}\right)=\mathbf{f}_{\mathrm{V}}\!\left(\mathbf{X}_{\mathrm{A}\rightarrow\mathrm{V}}\right)$. 

To acquire $\mathbf{X}_{\mathrm{A}\rightarrow\mathrm{V}}$, following the given intuition, we apply a generator $\mathbf{G}_{\mathrm{A}\rightarrow\mathrm{V}}$ and a segmentation model $\mathbf{f}_\mathrm{V}$ for phase $\mathrm{V}$, and design a min-max strategy to optimize them simultaneously. The objective is:
\begin{align}
\label{Eqn:Translation}
\begin{split}
\mathop{\mathrm{min}}_{\boldsymbol{\theta}_\mathrm{V}} \mathop{\mathrm{max}}_{\boldsymbol{\phi}_{\mathrm{A}\rightarrow\mathrm{V}}} \  & \mathbf{G}_{\mathrm{A}\rightarrow\mathrm{V}}\!\left(\mathbf{X}_{\mathrm{A}};\boldsymbol{\phi}_{\mathrm{A}\rightarrow\mathrm{V}}\right) \\
&+ \left\| \mathbf{f}_\mathrm{V}\!\left(\mathbf{G}_{\mathrm{A}\rightarrow\mathrm{V}}\!\left(\mathbf{X}_{\mathrm{A}};\boldsymbol{\phi}_{\mathrm{A}\rightarrow\mathrm{V}}\right);\boldsymbol{\theta}_\mathrm{V}\right) - \mathbf{Y}_{\mathrm{A}}\right\|
\end{split}
\end{align}
where $\boldsymbol{\theta}_\mathrm{V}$ and $\boldsymbol{\phi}_{\mathrm{A}\rightarrow\mathrm{V}}$ are the parameter of the segmentation model and the generator, respectively. The loss function thus involves a generation loss of $\mathbf{G}_{\mathrm{A}\rightarrow\mathrm{V}}$,  and a segmentation loss using the generated data $\mathbf{X}_{\mathrm{A}\rightarrow\mathrm{V}}$: 
\begin{align}
\label{Eqn:TransferLoss}
\begin{split}
&\mathcal{L}_\text{P2P}(\mathbf{X}_{\mathrm{A}},\mathbf{X}_{\mathrm{V}},\mathbf{Y}_{\mathrm{A}};\mathbf{G}_{\mathrm{A}\rightarrow\mathrm{V}}, \mathbf{f}_{\mathrm{V}}) \\
&= d\!\left[ \mathbf{G}_{\mathrm{A}\rightarrow\mathrm{V}}\!\left(\mathbf{X}_{\mathrm{A}};\boldsymbol{\phi}_{\mathrm{A}\rightarrow\mathrm{V}}\right),\mathbf{X}_{\mathrm{V}} \right]  \\
&+ \left\| \mathbf{f}_\mathrm{V} \left( \mathbf{G}_{\mathrm{A}\rightarrow\mathrm{V}}\!\left(\mathbf{X}_{\mathrm{A}};\boldsymbol{\phi}_{\mathrm{A}\rightarrow\mathrm{V}}\right);\boldsymbol{\theta}_\mathrm{V}\right) - \mathbf{Y}_{\mathrm{A}} \right\|
\end{split}
\end{align}
where $d$ indicates the similarity distance of the generated data $\mathbf{G}_{\mathrm{A}\rightarrow\mathrm{V}}\!\left(\mathbf{X}_{\mathrm{A}};\boldsymbol{\theta}_\mathrm{V}\right)$ and the true data $\mathbf{X}_{\mathrm{V}}$ in phase $\mathrm{V}$. Note that this distance can be measured with various methods, such as utilizing a discriminator as GAN\cite{goodfellow2014generative}, applying the distance like KL divergence \cite{joyce2011kullback} between the distribution $p(\mathbf{X}_{\mathrm{V}})$ and $p(\mathbf{X}_{\mathrm{A}\rightarrow\mathrm{V}}|\mathbf{X}_{\mathrm{A}})$, \emph{etc.}

\vspace{0.2cm}
\noindent
$\bullet$\quad{\bf The Overall Objective}

Combining Eqns~\eqref{Eqn:SegmentationLoss} and~\eqref{Eqn:TransferLoss}, the loss function of phase $\mathrm{A}$ is written as:
\begin{align}
\nonumber
\begin{split}
&\mathcal{L}_\mathrm{A}\!\left(\mathbf{X}_{\mathrm{A}},\mathbf{Y}_\mathrm{A},\mathbf{X}_{\mathrm{V}};\mathbf{f}_\mathrm{A},\mathbf{f}_\mathrm{V},\mathbf{G}_{\mathrm{A}\rightarrow\mathrm{V}}\right) \\
&= \!\left\|\mathbf{f}_\mathrm{A}\!\left(\mathbf{X}_\mathrm{A};\boldsymbol{\theta}_\mathrm{A}\right)-\mathbf{Y}_\mathrm{A}\right\| +  d\!\left[ \mathbf{G}_{\mathrm{A}\rightarrow\mathrm{V}}\!\left(\mathbf{X}_{\mathrm{A}};\boldsymbol{\phi}_{\mathrm{A}\rightarrow\mathrm{V}}\right),\mathbf{X}_{\mathrm{V}} \right]  \\
&+ \left\| \mathbf{f}_\mathrm{V} \left( \mathbf{G}_{\mathrm{A}\rightarrow\mathrm{V}}\!\left(\mathbf{X}_{\mathrm{A}};\boldsymbol{\phi}_{\mathrm{A}\rightarrow\mathrm{V}}\right);\boldsymbol{\theta}_\mathrm{V}\right) - \mathbf{Y}_{\mathrm{A}} \right\|
\end{split}
\end{align}
which is also shown as the red routine in Figure~\ref{Fig:Framework}. Note that the relation between phase $\mathrm{A}$ and $\mathrm{V}$ is symmetric, thus we apply another generator $\mathbf{G}_{\mathrm{V}\rightarrow\mathrm{A}}$ to transfer image data from $\mathrm{V}$ to $\mathrm{A}$. Now considering both phases $\mathrm{A}$ and $\mathrm{V}$, as in Figure~\ref{Fig:Framework}, yields the overall loss function for the proposed Phase Collaborative Network (PCN):
\begin{align}
\label{Eqn:OverallLoss}
\begin{split}
&\mathcal{L}_\mathrm{PCN}\!\left(\mathbf{X}_{\mathrm{A}},\mathbf{Y}_\mathrm{A},\mathbf{X}_{\mathrm{V}},\mathbf{Y}_{\mathrm{V}};\mathbf{S}_\mathrm{A},\mathbf{S}_\mathrm{V},\mathbf{G}_{\mathrm{A}\rightarrow\mathrm{V}},\mathbf{G}_{\mathrm{V}\rightarrow\mathrm{A}}\right) \\
&= \lambda \!\left\|\mathbf{f}_\mathrm{A}\!\left(\mathbf{X}_\mathrm{A};\boldsymbol{\theta}_\mathrm{A}\right)-\mathbf{Y}_\mathrm{A}\right\| +  d\!\left[ \mathbf{G}_{\mathrm{A}\rightarrow\mathrm{V}}\!\left(\mathbf{X}_{\mathrm{A}};\boldsymbol{\phi}_{\mathrm{A}\rightarrow\mathrm{V}}\right),\mathbf{X}_{\mathrm{V}} \right]  \\
&+ (1-\lambda) \left\| \mathbf{f}_\mathrm{V} \left( \mathbf{G}_{\mathrm{A}\rightarrow\mathrm{V}}\!\left(\mathbf{X}_{\mathrm{A}};\boldsymbol{\phi}_{\mathrm{A}\rightarrow\mathrm{V}}\right);\boldsymbol{\theta}_\mathrm{V}\right) - \mathbf{Y}_{\mathrm{A}} \right\| \\
&+ \lambda \!\left\|\mathbf{f}_\mathrm{V}\!\left(\mathbf{X}_\mathrm{V};\boldsymbol{\theta}_\mathrm{V}\right)-\mathbf{Y}_\mathrm{V}\right\| +  d\!\left[ \mathbf{G}_{\mathrm{V}\rightarrow\mathrm{A}}\!\left(\mathbf{X}_{\mathrm{V}};\boldsymbol{\phi}_{\mathrm{V}\rightarrow\mathrm{A}}\right),\mathbf{X}_{\mathrm{A}} \right]  \\
&+ (1-\lambda) \left\| \mathbf{f}_\mathrm{V} \left( \mathbf{G}_{\mathrm{V}\rightarrow\mathrm{A}}\!\left(\mathbf{X}_{\mathrm{V}};\boldsymbol{\phi}_{\mathrm{V}\rightarrow\mathrm{A}}\right);\boldsymbol{\theta}_\mathrm{V}\right) - \mathbf{Y}_{\mathrm{V}} \right\|
\end{split}
\end{align}
Note that here we introduce a coefficient $\lambda$ to adjust the weight of original and generated data, which ensures the stability and robustness of PCN in some cases.

\subsection{Implementation and Optimization}

We instantiate PCN using two state-of-the-art approaches, namely, RSTN~\cite{yu2018recurrent} and GAN~\cite{goodfellow2014generative}. RSTN is a coarse-to-fine segmentation model that consists of two jointly optimized segmentation networks. GAN is an unsupervised generative model that approximates the target distribution. Therefore, in Eqn~\eqref{Eqn:OverallLoss} $\mathbf{f}_\mathrm{A}$ and $\mathbf{f}_\mathrm{V}$ are applied with RSTN. Meanwhile, a pair of GANs consist of a translation model to calculate $d\!\left[ \mathbf{G}_{\mathrm{A}\rightarrow\mathrm{V}}\!\left(\mathbf{X}_{\mathrm{A}};\boldsymbol{\phi}_{\mathrm{A}\rightarrow\mathrm{V}}\right),\mathbf{X}_{\mathrm{V}} \right] + d\!\left[ \mathbf{G}_{\mathrm{V}\rightarrow\mathrm{A}}\!\left(\mathbf{X}_{\mathrm{V}};\boldsymbol{\phi}_{\mathrm{V}\rightarrow\mathrm{A}}\right),\mathbf{X}_{\mathrm{A}} \right]$. It is also remarkable that cycle consistency loss~\cite{CycleGAN2017} can be add in to generative loss to reinforce the generators, since the phase $\mathrm{A}$ and $\mathrm{V}$ are symmetric. In the experiment, for fair comparison, same architecture described in~\cite{yu2018recurrent} is applied for RSTN models, where coarse and fine models are both a FCN-8s model; the generators applied the same architecture described in \cite{CycleGAN2017}, where the generator has $9$ residual blocks and the discriminator is a $70\times70$ PatchGAN \cite{isola2017image}. 

Since both RSTN and GAN reported training instability to some extent, we train PCN, \textit{i.e.} equivalently optimize the above loss function, using a two-stage strategy, in which the segmentation and translation modules are first separately trained to guarantee convergence and then jointly optimized towards higher performance. This prevents the not-yet-well-trained segmentation model from outputting low-accuracy results which mislead the generative models, and vice versa.

\begin{itemize}
\item During the {\bf separate stage}, we set ${\lambda}={1}$ in Eqn~\eqref{Eqn:SegmentationLoss} so that each segmentation model, either in phase $\mathrm{A}$ or $\mathrm{V}$, is only trained with real data $\left(\mathbf{X}_{\mathrm{A}},\mathbf{Y}_{\mathrm{A}}\right)$ and $\left(\mathbf{X}_{\mathrm{V}},\mathbf{Y}_{\mathrm{V}}\right)$. At the meantime, the generators are trained in a way that is equivalent to the training of CycleGAN~\cite{CycleGAN2017}. Without this separate stage, the overall model often fails to converge.
\item During the {\bf joint stage}, we relax the constraints on $\lambda$ so that it falls within $\left(0,1\right)$. During this stage, segmentation and translation networks are optimized jointly, so that the segmentation model in each phase has some chances of receiving generated data from the other phase for training, and the translation model also `tunes' itself to generate more discriminative image data. Consequently, the qualities of both segmentation and translation become higher. We will show the necessity of this joint stage in the experimental section.
\end{itemize}

During testing, we simply follow the same flowchart as in training, and fuse the prediction in both phases as the final output. For simplicity, in this paper, the outputs of two phases are averaged when dealing with segmentation in either phase, \textit{i.e.} $\lambda = 0.5$, although tuning the fusing weight with a held-out validation dataset often leads to higher segmentation accuracy.

\subsection{Application to One-Phase Data}

Despite that PCN is formulated assuming two-phase data are available, it is easily applied to one-phase data. Let us take the NIH pancreas dataset~\cite{roth2015deeporgan} as an example, in which only the {\em venous} phase is collected and labeled. Here, the key is to borrow a well-trained translation model, from {\em venous} to {\em arterial}, learned from a two-phase dataset, and optimize it with the {\em venous} data. Note that, as in the two-phase scenario, the translation model can be fixed during training, making PCN degenerate to a model aware of unobserved {\em arterial} data generated from real {\em venous} data. For the same reason, better performance is achieved by joint learning. The testing process follows the same flowchart described previously, where segmentation results produced in both phases, one real and one generated, are averaged as the final prediction.

\renewcommand{\colwidth}{1.8cm}
\begin{table*}[!t]
\centering
\setlength{\tabcolsep}{0.08cm}
\begin{tabular}{|l|l||R{\colwidth}|R{\colwidth}|R{\colwidth}|R{\colwidth}|R{\colwidth}|R{\colwidth}|R{\colwidth}|}
\hline
                          & Organ  & \textit{adrenal g.} & \textit{gallbladder} & \textit{infer. v. c.} & \textit{kidney l.} & \textit{kidney r.} & \textit{pancreas} & \textit{super. m. a.} \\ \hline\hline
\multirow{2}{*}{Normal-A}   & RSTN-A & $59.40\%$             & $87.20\%$              & 73.62\%           & $94.28\%$            & $95.13\%$            & $84.58\%$           & $74.48\%$           \\ \cline{2-9}
                         & PCN-A & $\mathbf{64.96\%}$    & $\mathbf{87.58\%}$             & $\mathbf{77.09\%}$           & $\mathbf{94.44\%}$   & $\mathbf{95.81\%}$   & $\mathbf{84.89\%}$           & $\mathbf{79.30\%}$  \\ \hline \hline
\multirow{2}{*}{Normal-V}   & RSTN-V & $56.11\%$             & $87.19\%$             & $78.77\%$           & $\mathbf{94.26\%}$            & $92.10\%$            & $86.94\%$           & $71.67\%$           \\ \cline{2-9}
                          & PCN-V & $\mathbf{64.00\%}$             & $\mathbf{88.16\%}$     & $\mathbf{81.42\%}$  & $93.33\% $           & $\mathbf{95.49\%}$            & $\mathbf{88.20\%}$  & $\mathbf{74.36\%}$           \\ \hline\hline
\multirow{2}{*}{Abnormal-A} & RSTN-A & $58.14\%$             & $80.59\% $             & $73.20\%$           & $92.09\%$            & $94.45\%$            & $80.32\%$           & $66.28\%$           \\ \cline{2-9}
                          & PCN-A & $\mathbf{63.57\%}$    & $\mathbf{86.19\%}$              & $\mathbf{74.70\%}$           & $\mathbf{93.01\%}$            & $\mathbf{94.58\%}$   & $\mathbf{81.48\%}$  & $\mathbf{69.75\%}$  \\ \hline\hline
\multirow{2}{*}{Abnormal-V} & RSTN-V & $52.60\%$               & $86.10\%$              & $78.19\%$           & $92.78\%$            & $90.26\% $           & $75.89\%$          & $62.72\% $          \\ \cline{2-9}
                          & PCN-V & $\mathbf{59.90\%}$             &$ \mathbf{89.28\%}$     &$ \mathbf{79.43\%}$  &$ \mathbf{95.51\%}$   & $\mathbf{94.00\%}$           & $\mathbf{79.82\%}$           & $\mathbf{64.40\%}$           \\ \hline
\end{tabular}
\caption{DSC comparison between RSTN~\cite{yu2018recurrent} and PCN (our approach) on two-phase multi-organ segmentation.}
\label{Tab:TwoPhase}
\vspace{-0.5cm}
\end{table*}

\section{Experiments}
\label{Experiments}

\subsection{Datasets, Evaluation and Details}
\label{Experiments:Settings}
We collect a two-phase dataset that contains abdominal CT scans with multiple organs and blood vessels. There are $200$ normal and $200$ abnormal (with pancreatic cancer) patients in both the {\em arterial} and {\em venous} phases, respectively. We also refer to two public datasets which only contains single ({\em venous}) phase. The NIH pancreas segmentation dataset~\cite{roth2015deeporgan} contains $82$ CT scans, and the pancreas segmentation subset of the medical decathlon dataset (MSD)\footnote{{\sf http://medicaldecathlon.com/index.html}} contains $282$ CT scans, in all of which the pancreas was annotated. The resolution of each scan is $512\times512\times L$, where $L$ is the number of slices along the long axis of the body. The range of $L$ varies among these datasets, where on average $\bar{L}$ is $751$ for our data, $286$ for NIH data and $457$ for MSD data. All data are clamped into $\text{HU} \in [-125,275]$.

Following the conventions, we split each dataset into $4$ folds, each of which contains approximately the same number of samples. We train the models on $3$ out of $4$ subsets and test them on the held-out one. Different from single-phase baseline models, PCN is trained with data from two phases, while tested in a same setting as baseline models. We measure the segmentation accuracy by computing the Dice-S{\o}rensen coefficient (DSC)~\cite{Sorensen-1948-BK} for each sample, and report the average value over all cases. Training PCN with RSTN involves two sections. In the first section, we fix the translation module and and train RSTN with a learning rate of $10^{-5}$. $80\rm{,}000$ and $10\rm{,}000$ iterations are used in the separate and joint training stages of RSTN, respectively. In the second section, we train RSTN (the joint stage) together with the translation module for another $30\rm{,}000$ iterations. After every $10\rm{,}000$ iterations, we multiply the learning rate by a factor of $0.8$. It typically takes around $40$ GPU-hours to finish a complete PCN training process.

\subsection{Segmentation in Our Two-Phase Dataset}
\label{Experiments:TwoPhase}

\begin{figure}[!t]
\centering
\includegraphics[width=\linewidth]{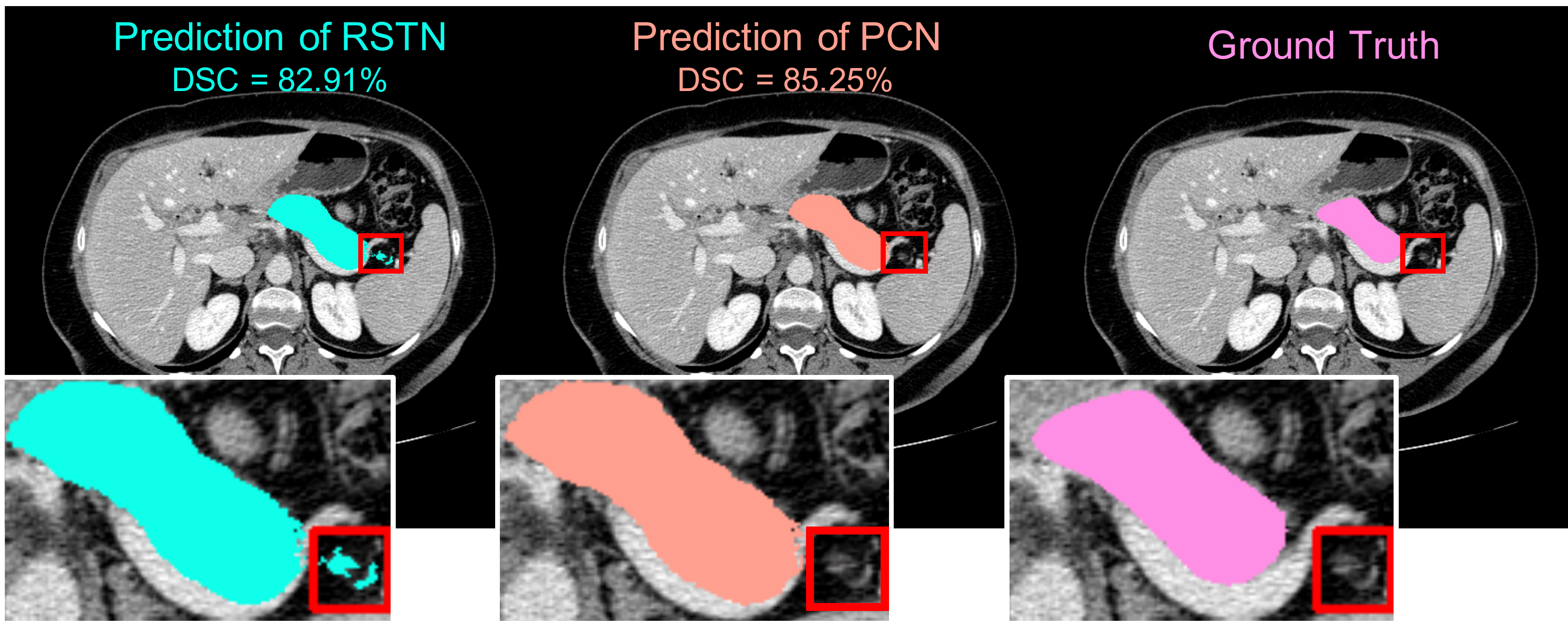}
\caption{(\textit{Best viewed in color}) Typical {\em pancreas} segmentation results, in the {\em axial} view, produced by RSTN and PCN, respectively. The red frames indicate a few false positive region that is eliminated by PCN.}
\label{Fig:Visualization}
\vspace{-0.5cm}
\end{figure}

\begin{figure*}[!t]
\centering
\includegraphics[width=\textwidth]{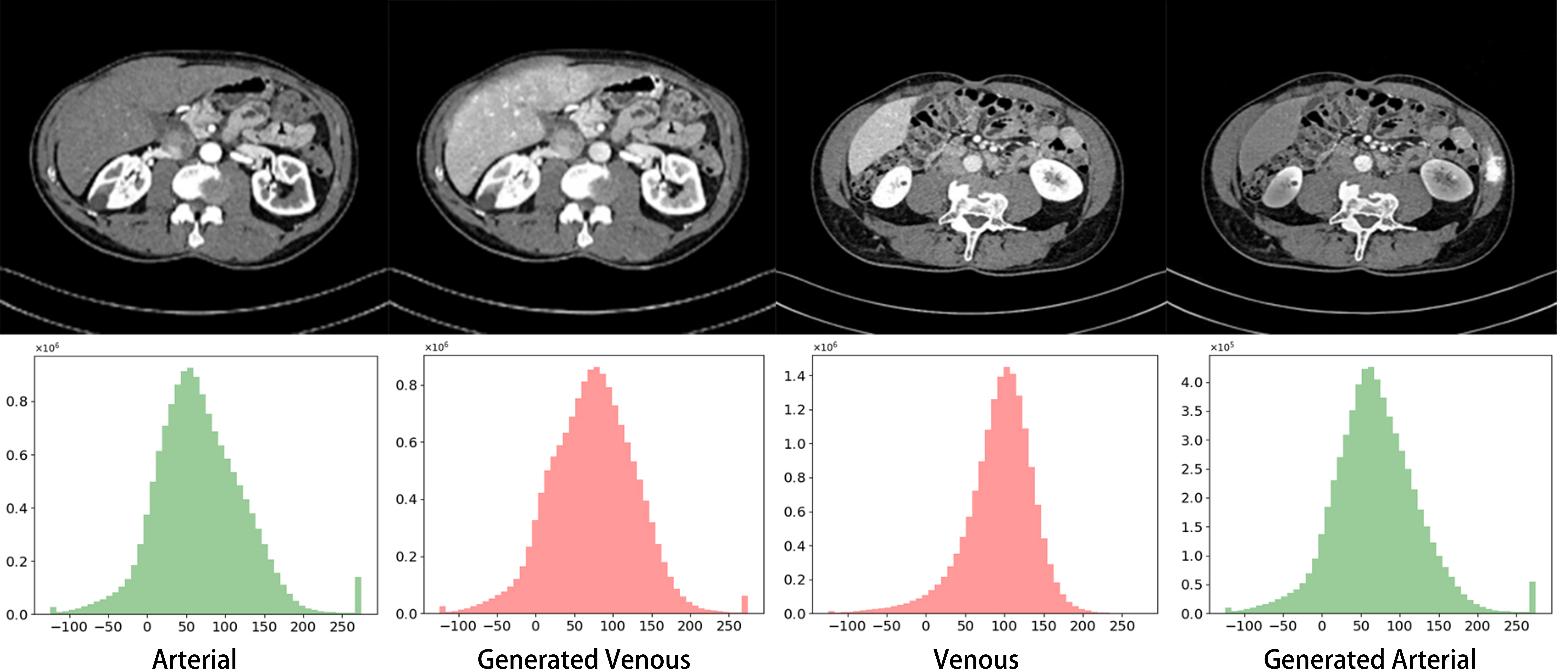}
\vspace{-0.5cm}
\caption{(\textit{Best viewed in color}) The top part shows the real and generated {\em arterial} and {\em venous} CT images, and the bottom part shows the distribution of intensity in the {\em pancreas} area (over all $200$ cases), determined by the ground-truth mask.}
\label{Fig:PhaseDistribution}
\vspace{-0.5cm}
\end{figure*}

We investigate seven targets in our dataset, which contains five abdominal organs and two blood vessels, one {\em artery} and one {vein}, which are better observed in the {\em arterial} and {\em venous} phases, respectively. Considering the normal and abnormal divisions, we have a total of $28$ tasks, each of which is trained and tested individually. Results are summarized in Table~\ref{Tab:TwoPhase}.

We first observe that different organs are better segmented in different phases. For example, the {\em superior mesenteric artery} and {\em inferior vena cava} prefer the {\em arterial} and {\em venous} phases, respectively, which is mainly caused by the properties in radiology. We also note a significant difference in segmenting the {\em gallbladder}, which reports comparable numbers in the normal cases but biases heavily towards the {\em venous} phase when {\em pancreas} abnormality is present -- this is partly due to the impact of pancreatic cancer, although the cancer makes the {\em pancreas} itself easier to be recognized in the {\em arterial} phase. All these results were verified reasonable by the radiologists in our team.

Then we use PCN to integrate two-phase information towards better segmentation. In $27$ out of $28$ individual tasks, PCN outperforms the corresponding single-phase baseline, with the only exception lying in the {\em venous} phase scan of normal {\em left kidney}, in which PCN is slightly outperformed by $1\%$. The most significant improvement is observed on the {\em adrenal gland}, a tiny target just above the {\em kidneys}, in which PCN outperforms the baseline by a large margin of at least $5\%$. This indicates that PCN indeed benefits from information fusion. We consulted the radiologists on the reason that, for example, on {\em inferior vena cava}, PCN majoring in the {\em arterial} phase works inferior to the baseline in the {\em venous} even after it sees {\em venous} information. The answer lies in that ground-truth annotation in the {\em arterial} phase is not as precise as that in the {\em venous} phase, because it is even difficult for the experts to distinguish the boundary of this minor {\em vein} in the {\em arterial} phase. Integrating two-phase information is the correct direction to work on.

Before we continue on single-phase datasets, we perform several diagnostic experiments to discuss on the behaviors of PCN on this two-phase dataset.

\vspace{0.2cm}
\noindent
$\bullet$\quad{\bf Qualitative Visualization}

We first show an example of {\em pancreas} segmentation in Figure~\ref{Fig:Visualization}. The single-phase ({\em venous}) baseline, RSTN~\cite{yu2018recurrent}, produces a false-positive area, and it is filtered out when complementary information provided by the {\em arterial} phase is integrated. Note that, unlike the blood vessels, it is not known which phase is better for {\em pancreas} segmentation (see Table~\ref{Tab:TwoPhase}). In this scenario, PCN is a safer choice which consistently improves segmentation accuracy in both phases.

\vspace{0.2cm}
\noindent
$\bullet$\quad{\bf Quality of Image Translation}

Next, we analyze the quality of image translation produced by the translation module. A typical example is shown in Figure~\ref{Fig:PhaseDistribution}, in which we can clearly observe the change of intensity in the {\em liver} (the large region at top-left, brighter in {\em venous}), as well as that in the main {\em artery} named {\em aorta} (the small round at the center, brighter in {\em arterial}). In addition, to provide a better view in statistics, we investigate the distributions of $p_\mathrm{data}\!\left(\mathbf{X}_\mathrm{A}\right)$ and $p_\mathrm{data}\!\left(\mathbf{X}_\mathrm{V}\right)$ which are also those considered in Equation \eqref{Eqn:Translation}. We make use of the ground-truth segmentation mask to obtain the intensity (HU) distributions of different organs on the real/generated {\em arterial}/{\em venous} data, and results are shown in Figure~\ref{Fig:PhaseDistribution}. We note that the peak of the HU distribution appears around $50$ and $100$ for real {\em arterial} and {\em venous} data, respectively, showing different distributions, and the generated {\em arterial} and {\em venous} data approximate these distributions very well.

\vspace{0.2cm}
\noindent
$\bullet$\quad{\bf Difference from Data Augmentation}

\renewcommand{\colwidth}{1.7cm}
\begin{table}[!t]
\centering
\setlength{\tabcolsep}{0.08cm}
\begin{tabular}{|l||R{\colwidth}|R{\colwidth}|R{\colwidth}|}
\hline
Organs                     & {\em infer. v. c.} & {\em pancreas} & {\em super. m. a.} \\ \hline\hline
{\em arterial} ($100$)     & $67.52\%$          & $79.29\%$          & $68.42\%$          \\ \hline
{\em venous} ($100$)       & $70.89\%$          & $78.32\%$          & $64.57\%$          \\ \hline
{\em mixed} ($50\times2$)  & $67.30\%$          & $74.45\%$          & $65.44\%$          \\ \hline
{\em mixed} ($200\times2$) & $\mathbf{74.45\%}$ & $\mathbf{81.23\%}$ & $\mathbf{72.69\%}$ \\ \hline
\end{tabular}
\caption{DSC comparison among RSTN models trained on different numbers (in parentheses) of training data.
\vspace{-0.5cm}
\noindent}
\label{Tab:DataAugmentation}
\end{table}

We present and discuss on an interesting question, namely, what is the difference between PCN and a data augmentation method that simply integrates two-phase data in the training stage? To answer it, we make a comparison among three experiments. The first one involves training two RSTN models on $100$ {\em arterial} and $100$ {\em venous} cases, respectively; the second one trains an RSTN using $50$ cases from each phase (a total of $100$ cases); and the third one uses all these data ($200$ cases) to train another RSTN. The testing set remains the same as in the main experiments (Table~\ref{Tab:TwoPhase}). Results are listed in Table~\ref{Tab:DataAugmentation}. Compared with Table~\ref{Tab:TwoPhase}, we can see that segmentation accuracy is not necessarily improved by (i) using both {\em arterial} and {\em venous} data in training or (ii) increasing the size of the training set. This is to say, the improvement brought by PCN not only comes from seeing more training data, but also from the ability of modeling the relationship between different phases.

\vspace{0.2cm}
\noindent
$\bullet$\quad{\bf Ablation Study: Unsupervised Domain Adaptation}

Last but not least, we study the impact of four training strategies to demonstrate the necessity of the min-max strategy mentioned in Equation~\eqref{Eqn:Translation}. (i) {\bf PCN-2}: the standard PCN on a two-phase dataset; (ii) {\bf PCN-1}: PCN on a single-phase dataset (in this setting, we assume that training data on another phase are not observed, and thus we apply a pre-trained translation module and fix it during training); (iii) {\bf UDA-2}: the translation module is trained in an Unsupervised Domain Adaptation (UDA) manner, which means we keep $\lambda=1$ in Equation \eqref{Eqn:OverallLoss}. (iv) {\bf UDA-1}: PCN with a fixed and UDA-pretrained translation module; here we applied a CycleGAN. Results on two blood vessels (on which we expect two-phase information to play very important roles) are summarized in Table~\ref{table:training config}. With either module fixed, we observe consistent accuracy drop, which implies the necessity of our motivation, {\em i.e.}, modeling both data-to-label relation $p_\mathrm{data}\!\left(\mathbf{Y}_\mathrm{A}\mid\mathbf{X}_\mathrm{A}\right)$ and the phase-to-phase relation $p_\mathrm{data}\!\left(\mathbf{X}_\mathrm{V}\mid\mathbf{X}_\mathrm{A},\mathbf{Y}_\mathrm{A}\right)$, which are jointly accomplished by the segmentation and translation modules.

A side note is that, making segmentation and translation modules independent, we are mimicking the behavior of CyCADA~\cite{Hoffman_cycada2017} and SIFA \cite{chen2019synergistic}, segmentation approaches with an unsupervised domain adaptation translation module. It is thus a degenerated version of PCN, and reports lower accuracy than PCN in our setting that phase-to-phase relation is critical for segmentation.

\newcommand{\colwidthA}{1.5cm}
\newcommand{\colwidthB}{1.7cm}
\begin{table}[tb]
\centering
\setlength{\tabcolsep}{0.08cm}
\begin{tabular}{|l||R{\colwidthA}|R{\colwidthB}||R{\colwidthA}|R{\colwidthB}|}
\hline
Phase         & \multicolumn{2}{c||}{{\em arterial}}   & \multicolumn{2}{c|}{{\em venous}}     \\ \hline
Organ         & {\em infer. v. c.} & {\em super. m. a.} & {\em infer. v. c} & {\em super. m. a.} \\ \hline\hline
PCN-2     & $\mathbf{77.09\%}$ & $\mathbf{79.30\%}$ & $\mathbf{81.42\%}$ & $\mathbf{74.36\%}$ \\ \hline
PCN-1    & $74.32\%$          & $75.28\%$          & $78.96\%$          & $73.62\%$          \\ \hline
UDA-2     & $73.85\%$          & $74.43\%$          & $78.80\%$          & $73.14\%$          \\ \hline
UDA-1    & $73.14\%$          & $72.01\%$          & $72.66\%$          & $68.11\%$          \\ \hline
\end{tabular}
\caption{Segmentation accuracy (DSC) comparison among different training configurations (see texts for details).}
\label{table:training config}
\vspace{-0.5cm}
\end{table}

\subsection{Segmentation in Single-Phase Datasets}
\label{Experiments:OnePhase}

Most public medical image datasets have only one phase, because collecting two-phase data with annotations is often expensive in both time and labor. We generalize PCN to this scenario to demonstrate its potential application, that uses the pre-learned phase-to-phase relation to assist single-phase segmentation. Both NIH and MSD-{\em pancreas} datasets are in the {\em venous} phase, so we directly apply a pre-trained {\em venous}-to-{\em arterial} generator to {\em arterial} image generation. The parameters of this generator are fixed, {\em i.e.}, we only train the segmentation models during optimization.

Segmentation results are shown in Table~\ref{Tab:OnePhase}, which shows that, in term of average DSC, PCN outperforms its direct baseline, RSTN, consistently. Note that the accuracy of $84.50\%$ reported by RSTN is already the state-of-the-art in the NIH dataset, and PCN still outperforms RSTN in $74$ out of $82$ testing cases. In particular, we note that the worst case in both datasets are largely boosted by PCN, which further suggests that complementary information is provided by another ({\em arterial}) phase, and such information is especially useful for the most difficult cases which, according to the radiologists, are mostly caused by some extents of missing information in the given ({\em venous}) phase.

To investigate how PCN benefits from the {\em arterial} phase, we study network predictions on the generated input images. An average statistics shows the segmentation network gain $553$ out of $2\rm{,}258$ and $929$ out of $4\rm{,}224$ correct prediction pixels from {\em arterial} information on NIH and MSD-{\em pancreas} dataset respectively. Also considering the effect of filtering false positives, as shown in Figure~\ref{Fig:Visualization}, the complementariness of data from another phase becomes clear.

\begin{table}[tb]
\centering
\setlength{\tabcolsep}{0.12cm}
\begin{tabular}{|l|l||r|r|r|}
\hline
Data                 & Approach    & Average & Max     & Min     \\ \hline
\multirow{4}{*}{NIH} & Roth {\em et al.}~\cite{roth2015deeporgan} & $78.01\%$ & $88.65\%$ & $34.11\%$ \\ \cline{2-5} 
                     & Zhou {\em et al.}~\cite{zhou2017fixed}     & $82.37\%$ & $90.85\%$ & $62.43\%$ \\ \cline{2-5} 
                     & Yu {\em et al.}~\cite{yu2018recurrent}     & $84.50\%$ & $91.02\%$ & $62.81\%$ \\ \cline{2-5} 
                     & Ours (PCN)                                 & $\mathbf{85.15\%}$ & $\mathbf{94.68\%}$ & $\mathbf{68.89\%}$ \\ \hline\hline
\multirow{2}{*}{MSD} & Yu {\em et al.}~\cite{yu2018recurrent}     & $73.38\%$ & $\mathbf{87.54\%}$ & $35.53\%$ \\ \cline{2-5} 
                     & Ours (PCN)                                 & $\mathbf{76.59\%}$ & $86.23\%$ & $\mathbf{57.29\%}$ \\ \hline
\end{tabular}
\caption{Segmentation accuracy (DSC) comparison between our approach and the state-of-the-arts on the NIH and MSD datasets for {\em pancreas} segmentation.}
\label{Tab:OnePhase}
\vspace{-0.5cm}
\end{table}

\section{Conclusions}
\label{Conclusions}

In this paper, we present the Phase Collaborative Network (PCN) to deal with two-phase segmentation in the area of medical image analysis. From a theoretical perspective, we demonstrate that the difficulty mainly lies in modeling (i) the gap between images from different views and (ii) the fact that both labels, though not pixel-wise aligned, are sampled from the same distribution. We study this problem by modeling two relations, namely, data-to-label relation and phase-to-phase relation, for which we propose to combine the segmentation module with a generative module. The entire network is optimized in an end-to-end manner, and achieves satisfying performance on both single-phase and two-phase datasets. Confirmed by the radiologists in our team, these segmentation results are helpful to computer-assisted clinical diagnoses.

The success of our approach lays the foundation of integrating multi-phase data into various vision problems. However, as a preliminary study, PCN still suffers some drawbacks. For example, we make use of a GAN-based method to model phase-to-phase relation, but such models can be heavily constrained by the domains in training data, which limits its application to other types of data, {\em e.g.}, a generative model pre-trained in abdominal CT scans is unlikely to transfer to brain CT scans or even MRI scans. Also, PCN assumes that both phases contain part of known information, but in real-world, it is possible that image data in both phase are available but annotations are provided only in one phase. In this scenario, a possible direction is to integrate PCN with weakly-supervised or semi-supervised approaches, which are left for future research.

\section*{Acknowledgements}
This paper was supported by the Lustgarten Foundation for Pancreatic Cancer Research. We thank Qing Liu and Zihao Xiao for enormous help in improving the quality of this paper. We also appreciate Wei Shen, Seyoun Park, Yan Wang, Yuyin Zhou, Chenxi Liu, Zhuotun Zhu, Yingda Xia, Fengze Liu, Jiangchao Yao, Xu Chen, Jie Chang, Maosen Li and Jialiang Lu for instructive discussions.

\newpage

{\small
\bibliographystyle{ieee}
\bibliography{egbib}
}

\end{document}